    \pdfoutput=1

    \documentclass[11pt, a4paper]{article}

    \usepackage[table]{xcolor}
    \usepackage[backref=page]{hyperref}

    \usepackage{acl}
    
    \usepackage{times}
    \usepackage{latexsym}
    
    \usepackage[utf8]{inputenc}
    
    \usepackage{enumitem}
    
    \usepackage{geometry} 
    
    \usepackage[T1]{fontenc}
    \usepackage{footnotebackref}
    
    \usepackage{subfiles}

    \usepackage{amsmath}        
    \usepackage{amssymb}
    \usepackage{amsfonts}       
    \usepackage{amsthm}         
    \usepackage{bbding}         
    \usepackage{bm}             
    \usepackage{bbm}             
    \usepackage{graphicx}       
    \usepackage{fancyvrb}       
    \usepackage{dcolumn}        
    \usepackage{booktabs}       
    \usepackage{pifont}

    \usepackage{microtype}
    \usepackage{listings}
    \usepackage{mathtools}
    \usepackage{titlesec}
    \usepackage{float}
    \usepackage[frozencache,cachedir=.]{minted}
    \usepackage{multirow}

    \usepackage{caption}
    \usepackage{fancyhdr}
    \usepackage{nicefrac}

    \usepackage{url}

    \usepackage[nameinlink,capitalize,noabbrev]{cleveref}

    \makeatletter
    \newcommand\crulefill{\leavevmode
    \begingroup 
    \setlength{\dimen@}{0.5ex}
    \addtolength{\dimen@}{0.4pt}
    \leaders\hrule height\dimen@ depth -0.5ex \hfill
    \endgroup
    \kern\z@}
    
\newif\ifbackrefshowonlyfirst
\backrefshowonlyfirstfalse
\makeatletter
\let\BR@direct@old@hyper@natlinkstart\hyper@natlinkstart
\renewcommand*{\hyper@natlinkstart}{\phantomsection\BR@direct@old@hyper@natlinkstart}
\let\BR@direct@oldBR@citex\BR@citex
\renewcommand*{\BR@citex}{\phantomsection\BR@direct@oldBR@citex}%

\long\def\hyper@page@BR@direct@ref#1#2#3{\hyperlink{#3}{#1}}

\ifx\backrefxxx\hyper@page@backref
    \let\backrefxxx\hyper@page@BR@direct@ref
    \ifbackrefshowonlyfirst
    \fi
\else
    \ifbackrefshowonlyfirst
    \fi
\fi

\RequirePackage{etoolbox}
\patchcmd{\Hy@backout}{Doc-Start}{\@currentHref}{}{\errmessage{I can't seem to patch backref}}
\makeatother

    \title{Trained on 100 million words and still in shape: \\BERT meets British National Corpus}

    \author{David Samuel, Andrey Kutuzov, Lilja Øvrelid and Erik Velldal \\
      University of Oslo, Language Technology Group \\
      \texttt{\{davisamu, andreku, liljao, erikve\}@ifi.uio.no} \\}

\begin{document}
    \maketitle
    
    \begin{abstract}
    
While modern masked language models (LMs) are trained on ever larger corpora, we here explore the effects of down-scaling training to a modestly-sized but representative, well-balanced, and publicly available English text source -- the British National Corpus. We show that pre-training on this carefully curated corpus can reach better performance than the original BERT model. We argue that this type of corpora has great potential as a language modeling benchmark. To showcase this potential, we present fair, reproducible and data-efficient comparative studies of LMs, in which we evaluate several training objectives and model architectures and replicate previous empirical results in a systematic way. We propose an optimized LM architecture called LTG-BERT.
    
    \end{abstract}
    
    \section{Introduction}
    In the pursuit of state-of-the-art performance, NLP practitioners utilize increasingly larger amounts of data to pre-train language models, 
    making it difficult to disentangle the improvements made by the proposed modeling choices themselves. 
    Instead, our aim is to shift the focus towards more efficient language modeling on a small and standardizable pre-training corpus.
    We study the data efficiency of current language models on an openly available corpus of approximately 100M words -- incidentally the estimated amount of words processed by humans before adulthood \citep{linzen-2020-accelerate}.

The goal of this paper is not to rival the paradigm of `massively pre-trained language models'; instead we would in this work like to pursue a complementary direction of language modeling, which will hopefully lead to more interest in data-efficient language models. In particular, our contribution in this paper is twofold -- we show that:

    \noindent
    \paragraph{1.}\textbf{100M words is enough} to train a competitive language model that outperforms the downstream performance of the original BERT model. We show that the combination of a well-curated representative corpus, improved LTG-BERT architecture and a better training objective results in a model with stronger linguistic knowledge than the original English BERT pre-trained on $30\times$ larger corpus.
    
     Large language models are notoriously data hungry, requiring hundreds of gigabytes of raw textual data. This becomes a major obstacle for low-resource languages while also putting a limit to the efficiency of any `efficient' language model. On top of that, the size of web-crawled corpora makes it almost impossible to control their content and to prevent learning from harmful or copyrighted text \citep{bender2021dangers}. The British National Corpus \citep[BNC;][]{bnc2007british} is a 100-million-word reference corpus, manually curated to cover most aspects of $20^\textrm{th}$ century British English. 
    
    \noindent
    \paragraph{2.}\textbf{Reproducibility and fair comparison} of language models can be easily achieved by pre-training on the 
    British National Corpus. 
    
    Massive language models are often pre-trained on nonpublic filtered collections of web-crawled text, which makes any reproduction impossible. 
    We pre-train our models on a small and publicly available corpus, 
    which allows for a replicable comparison of different language modeling configurations and which can be easily utilized in future research of novel variants of language models. We also release the pre-processing scripts, training scripts as well as the final model checkpoints.\footnote{\url{https://github.com/ltgoslo/ltg-bert}}
    
    Previously, language models have been pre-trained on different corpora tokenized by different tokenizers and fine-tuned by increasingly complex learning methods. This makes any comparison of the underlying neural architectures and pre-training objectives unfair. We make the language models in this paper directly comparable by fixing the training corpus, the tokenizer and the evaluation methods, while keeping them as simple as possible.

\section{Related Work}

The data requirements of language models have been growing in orders of magnitude since their early stages \citep{jelinek1976}. Taking a huge leap towards more recent work, 
ELMo \citep[Embeddings from Language Models; ][]{peters-etal-2018-deep} were the first to introduce deep \textit{contextualized} embeddings of words. Recognizing the need of a large text corpus for this task, ELMo was trained on the 1B Word Benchmark \citep{chelba14_interspeech}. Later, BERT \citep[Bidirectional Encoder Representations from Transformers; ][]{devlin-etal-2019-bert} further advanced the performance of contextualized embeddings when it based the entire language model on the Transformer architecture \citep{NIPS2017_3f5ee243}. Another important aspect of BERT is that it was trained on a larger corpus than ELMo: about 3.3B words from crawled English Wikipedia and BookCorpus \citep{7410368}. To our best knowledge, the exact version of neither of the two subcorpora is publicly available.\footnote{BookCorpus \citep{7410368} is not available anymore and the authors of BERT do not specify what version of Wikipedia dump they used or how did they preprocess it (\url{https://github.com/google-research/bert\#pre-training-data}).} The issue of limited replicability has become even more pronounced with later large language models: XLNet \cite{NEURIPS2019_dc6a7e65} was trained on 33B words, RoBERTa \citep{DBLP:journals/corr/abs-1907-11692} on more than 30B words and GPT-3 \citep{NEURIPS2020_1457c0d6} on an approximately 400B word corpus. None of these datasets is available; the authors utilize non-trivial filtering algorithms but do not release the end product nor the filtering scripts.

The benefits of large corpora were questioned in CamemBERT \citep{martin-etal-2020-camembert} and the effect of corpus size has been then thoroughly studied in \newcite{micheli-etal-2020-importance}, \newcite{zhang-etal-2021-need} as well as in \newcite{https://doi.org/10.48550/arxiv.2203.15556}. They test differently sized random subsets of a BERT-like corpus (crawled Wikipedia and Smashwords) and of a massive web-crawled text corpus \citep[MassiveText;][]{https://doi.org/10.48550/arxiv.2112.11446}, respectively. Unlike them, we evaluate the effect of training on a small corpus, which was \textit{carefully curated} to create a representative sample of English. The British National Corpus is arguably more diverse and informative than a random subset of a web crawl -- hence we test how the \textit{quality} of a pre-training corpus influences the downstream performance, not only how the data quantity matters. We believe this aspect is vital for the future research of effective and reliable language models.

{\renewcommand{\arraystretch}{1.2}
\begin{table}[t]
    \resizebox{\columnwidth}{!}{
        \begin{tabular}{@{}lrrrr@{}}
        \toprule
        \textbf{}   & \hspace{-3em}\textbf{documents} & \textbf{sentences} & \textbf{words} & \textbf{subwords} \\ \midrule
        train       & 4\,014           & 8\,501\,376            & 115\,870\,549      & 131\,392\,103         \\
        development & 35               & 106\,566             & 1\,215\,306        & 1\,367\,570           \\ \bottomrule
        \end{tabular}
    }
    \caption{Size of the train-development splits for the preprocessed BNC corpus. Note that the number of words is larger than the 100 million reported by the BNC Consortium due to our less conservative pre-tokenization strategy.}
    \label{tab:bnc}
\end{table}
}
 
    \section{British National Corpus}
    \label{sec:bnc}

    We use the British National Corpus (BNC) as a diverse, balanced, compact, and publicly available monolingual English corpus. 
    BNC is comprised of both written and spoken language with a total of 100 million words. 
    The manually curated content contains a wide range of British English from the late $20^\textrm{th}$ century -- newspapers, journals, books (academic and fiction), letters, essays, unscripted informal conversations or transcribed business meetings, radio shows or phone calls. 
    The written part makes up approximately 90\% of the corpus and the remaining 10\% contains the transcribed speech.
    The sources are truncated to contain at most 45\,000 words to ensure greater diversity within the limited amount of 100 million words.

    \paragraph{Creation.} The process of creating the BNC is extensively described in its documentation on the website.\footnote{\url{https://ota.bodleian.ox.ac.uk/repository/xmlui/handle/20.500.12024/2554}} It was created by the so called `BNC Consortium' led by Oxford University Press, and including major dictionary publishers Longman and Larousse Kingfisher Chambers; academic research centres at Oxford University Computing Services, the University Centre for Computer Corpus Research on Language (UCREL) at Lancaster University, and the British Library's Research and Innovation Centre. The purpose of the British National Corpus project was to construct a balanced and representative sample of current British English at the time. It was created over a period of four years and was a result of careful planning and data selection across a number of selection criteria (domain, time, medium, level) with proportions in the corpus designed to reflect the proportions found in real language use. It is widely acknowledged that the BNC has been a major influence on the construction of language corpora \citep{WheredidweGoWrongARetrospectiveLookattheBritishNationalCorpus}. One downside of the BNC is that it does not reflect anything occurring to English language and the world in the 21st century, but still no better alternatives of the same size and quality exists. In addition, BNC was used as a model for creating representative corpora for other languages: e.g., Turkish \citep{aksan-etal-2012-construction}.

    \paragraph{Version.} We use the third release of the corpus, BNC XML Edition (2007), which is the final revision of the texts compiled from 1991 to 1994 \citep{bnc2007british}.
    The XML edition did not get any additional content on top of the original text samples, but it got some minor corrections, more metadata and it is supplied in a convenient XML format.

    \subsection{Preprocessing}
    \label{sec:preprocessing}
    
    We convert the XML version of BNC into the Markdown format,\footnote{\url{https://daringfireball.net/projects/markdown/}} to make it human-readable and usable as a direct raw-text input of a language model. On top of that, it can also preserve some meta-information encoded in the original XML format. Short samples from the preprocessed corpus can be found in \cref{sec:bnc-samples}. After preprocessing, the articles are randomly placed into a training split and a development split. The proportions of both splits are given in \cref{tab:bnc}.
    
    \paragraph{Composition.} BNC is hierarchically composed of the following text units: words, sentences, paragraphs and articles. We preserve the sentence information by storing each sentence on a separate line; paragraphs are divided by a blank line and an article always starts with a top-level header. The word-tokens are intentionally not preserved -- instead, we heuristically detokenize the text to move it towards the natural text distribution. BNC includes information about the original whitespace, but we found it unreliable in some cases, necessitating the use of heuristics.

    \paragraph{Other metadata.} Other meta information available in our Markdown version is as follows:
    \begin{enumerate}
        \item \textbf{Headers:} We keep the headers together with their level by converting them to the atx-style format prefixed by hash symbols `\texttt{\#}'.
        \item \textbf{Speakers:} The spoken part of BNC is divided into speech turns, each accompanied by a speaker identifier. We maintain this information by formatting each speech turn as `\texttt{\{name\}:\textvisiblespace'\{turn\}'}'.
        \item \textbf{Quotes:} Markdown also allows us to keep the special quoted text by using a prefix `\texttt{>\textvisiblespace}'.
        \item \textbf{Lists:} The XML format contains special tags for lists and their respective elements, we use the `\texttt{-\textvisiblespace\{element\}}' notation to encode these text blocks.
        \item \textbf{Incomprehensible speech:} Some words or phrases could not be transcribed because they were illegible or inaudible. Since completely omitting such text would result in ungrammatical sentences, we mark these segments with a special `\texttt{[UNK]}' token.
    \end{enumerate}
    
    \noindent
    Not all of this additional information is of use for the language models tested in this article, but it can be easily filtered out when needed. We preserve it to make this corpus more versatile.
    
    \section{Model architecture}
    \label{sec:architecture}

    \begin{figure}[t]
        \centering
        \includegraphics[width=\columnwidth]{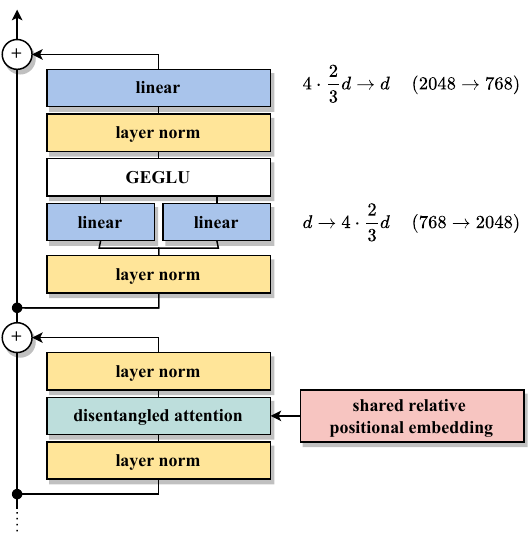}
        \caption{A simplified diagram of one layer in our LTG-BERT language model, which illustrates the changes made to the standard Transformer architecture -- NormFormer layer normalization, GEGLU activation function and disentangled attention.}
        \label{fig:architecture}
    \end{figure}
    
    We slightly depart from the typical \textit{post-norm} Transformer architecture \citep{NIPS2017_3f5ee243} used by BERT \citep{devlin-etal-2019-bert}, as illustrated in \cref{fig:architecture}. Preliminary experiments with this model showed that it tends to unpredictably diverge in the later stages of training. This behavior has been noted in previous work on large LMs \citep{liu-etal-2020-understanding} and accordingly, we follow some of the recent improvements of Transformer.
    
    \paragraph{NormFormer.} \textit{Pre-norm} variation of the Transformer has been shown to lead to more stable convergence with slightly degraded performance \citep{nguyen-salazar-2019-transformers}. \newcite{shleifer2022normformer} claimed to mitigate this degradation by introducing an additional layer normalization operation. For these reasons, we decided to use their so-called \textit{NormFormer} architecture to stabilize the training.\footnote{They also proposed some additional improvements -- \textit{head scaling} and \textit{residual scaling}, but we did not experience any performance benefits from these changes.}
    
    \paragraph{GEGLU activation function,} proposed in \newcite{DBLP:journals/corr/abs-2002-05202}, enhances the expressiveness of the original Transformer feed-forward modules by redefining them as
    $$\textrm{FF}_{\textrm{GEGLU}}(\bm{x}) = \left(\textrm{GELU}(\bm{xW}_1) \odot \bm{xW}_2\right)\bm{W}_3,$$ where $\bm{W}_i$ are weight matrices\footnote{The bias terms are omitted for brevity.} and GELU is the Gaussian Error Linear Unit \citep{Hendrycks2016}. Note that this formulation involves three linear transformations instead of two, we therefore lower the intermediate hidden size by $\nicefrac{2}{3}$ to keep the number of parameters the same.
    
    \paragraph{Disentangled attention.} The original Transformer formulation \citep{NIPS2017_3f5ee243} fuses the content and positional information together in the first embedding layer and calculates the (unnormalized) attention score between each pair of tokens $\bm{x}_i$ and $\bm{x}_j$ as
    $$\bm{A}_{i,j} = \frac{\bm{Q}_i\bm{K}_j^\intercal}{\sqrt{d}},$$
    where $\bm{Q}$ and $\bm{K}$ are the query-key linear transformations of $\bm{x}$.

    \newcite{he2021deberta} proposed to \textit{disentangle} the content and positional information. The content representations are incrementally built by the Transformer layers and the position is encoded by one shared relative positional embedding matrix $P \in \mathbb{R}^{(2L - 1) \times d}$, where $L$ is the maximal input length.\footnote{Tokens at positions $i$ and $j$ have relative positional embedding at the $(L - i + j)^\textrm{th}$ row of $P$, denoted as $P_{i,j}$.} This is supposed to offer greater expressivity as each layer can access these two parts directly. The attention scores are then calculated as a sum of three distinct parts: \textit{content-to-content}, \textit{content-to-position} and \textit{position-to-content} attention -- formally, the attention scores are defined as
    
    $$\bm{A}_{i,j} = \frac{\prescript{c\!}{}{\bm{Q}}_i\prescript{c\!}{}{\bm{K}}_j^\intercal + \prescript{c\!}{}{\bm{Q}}_i\prescript{p\!}{}{\bm{K}}_{i,j}^\intercal + \prescript{p\!}{}{\bm{Q}}_{j, i}\prescript{c\!}{}{\bm{K}}_j^{\intercal}}{\sqrt{3d}},$$
    where $\prescript{c\!}{}{\bm{Q}}$ and $\prescript{c\!}{}{\bm{K}}$ are linear transformations of the \textit{content} vectors and $\prescript{p\!}{}{\bm{Q}}$ and $\prescript{p\!}{}{\bm{K}}$ are linear transformations of the relative \textit{positional} embedding $P_{i,j}$. We share the parameters of the content and positional transformations, $\prescript{c\!}{}{\bm{Q}} = \prescript{p\!}{}{\bm{Q}}$ and $\prescript{c\!}{}{\bm{K}} = \prescript{p\!}{}{\bm{K}}$, to not increase the model size while achieving comparable performance \citep{he2021deberta}.

    \paragraph{Initialization scaling.} \newcite{https://doi.org/10.48550/arxiv.2204.06644} found that we can further stabilize the Transformer architecture by gradually scaling down its feed-forward (FF) weight matrices. Following \newcite{nguyen-salazar-2019-transformers}, we first initialize all weight matrices $\bm{W}$ by sampling from:
    $$\bm{W}_{i,j} \sim \mathcal{N}\!\left(0,\, \sqrt{\frac{2}{d + 4d}} \right),$$
    where $d$ is the hidden dimension.\footnote{This formula is roughly equal to the universal BERT initialization range of $0.02$ for $d=1024$.}
    Then all three weight matrices in a FF module at layer $l$ are scaled down by a factor of $\nicefrac{1}{\sqrt{2(l+1)}}$.

    \section{Training objectives}
    \label{sec:objectives}
    
    The fixed corpus, tokenizer and fine-tuning procedures establish a controlled test bed for a comparative study of training objectives proposed in the past. The original BERT model is trained via two self-supervised training objectives -- masked language modeling (MLM) and next sentence prediction (NSP). We evaluate five different configurations of these objectives (three for MLM and two for NSP), as further detailed below.  
    
    \subsection{Masked language modeling (MLM)}
    \label{sec:mlm}
    
    Unlike the traditional auto-regressive language models, the \textit{Bidirectional} Encoder Representations from Transformers (BERT) learn a \textit{bidirectional} contextualized representation for each token in a text segment. This is done by randomly selecting 15\% of subword tokens (excluding the special tokens). Out of these, 80\% are masked, 10\% randomly replaced and 10\% are left untouched. The language model is then trained to jointly predict the original state of the selected units. We investigate three common choices of the masked text units:
    
    \begin{enumerate}
        \item \textbf{Subwords}. As proposed in the seminal work by \newcite{devlin-etal-2019-bert}, every subword is masked independently with 15\% probability to model its bidirectional dependencies.

        \item \textbf{Whole words}. This method was also implemented by \newcite{devlin-etal-2019-bert}, after the publication of their original paper with subword masking. The motivation for this approach is that partially masked multi-subword word units are often easily decoded without any need for non-local contextual information; masking the whole multi-subword unit should force the model to build longer-range non-local dependencies. 
        
        \item \textbf{Spans}. The third method further follows the direction of whole-word masking and generalizes it to masking of random \textit{spans} of subwords. More specifically, SpanBERT \citep{joshi-etal-2020-spanbert} iteratively samples random spans until 15\% of subwords are masked. For each span, it first samples its length from $\textrm{Geo}(p)$, where $p=\nicefrac{1}{3}$.\footnote{To ensure that the sampled length is not too large, we take the sampled value modulo 10. The expected length of a masked span is then approximately equal to 2 with $p=\nicefrac{1}{3}$.} Then the starting subword of the masked span is chosen from a uniform distribution.
 \end{enumerate}
    
    \subsection{Next sentence prediction (NSP)}
    \label{sec:nsp}

    Masked language modeling is a token-level training objective that trains the model to learn rich token representations.
    Yet, some downstream tasks need a single sentence-level representation instead. To also learn these, researchers have designed a number of additional semi-supervised training objectives. On the other hand, \newcite{DBLP:journals/corr/abs-1907-11692} argue that NSP objectives do not help the downstream performance and they can thus be dropped in favour of a simpler optimization process with a single MLM training objective. To test these hypotheses, we experiment with two NSP objectives:
    
    \begin{enumerate}
        \item \textbf{Document discrimination.} \newcite{devlin-etal-2019-bert} sample two text segments and then train the model with a second discriminative loss function, which predicts whether the two segments are continual or randomly taken from two different documents.
        
        \item \textbf{Sentence-order discrimination.} \newcite{Lan2020ALBERT} argue that the document discrimination is too easy as the language models only have to compare the topic of the two segments to achieve a good performance in this task. Instead, they propose to predict whether the two segments are in the correct order or whether they are swapped. Thus, the sentence-order loss forces the neural network to model inter-sentence coherence and this is believed to lead to a better downstream performance.
    \end{enumerate}
    

    \section{Evaluation metrics}
    \label{sec:metrics}
    
    We use three conceptually different methods for evaluating the amount of linguistic knowledge acquired by the BNC language models. 1) The (Super)GLUE datasets test the ability of the model to adapt to various NLU tasks by further optimizing the whole pre-trained model, 2) edge probing tasks evaluate how much linguistic information one can extract from a frozen pre-trained model and 3) BLiMP utilizes the intrinsic ability of the pre-trained network to model language and probes its knowledge without any additional training. We further elaborate on each of these below. 
    
    \subsection{(Super)GLUE}

    GLUE \citep{wang-etal-2018-glue} and SuperGLUE \citep{NEURIPS2019_4496bf24} have become a de-facto standard for evaluating the language understanding capabilities of language models. Accordingly, we also choose to fine-tune our language models on these NLU tasks to measure their linguistic and transfer-learning performance. We give more technical details about our implementation of (Super)GLUE fine-tuning in \cref{sec:superglue-details}.

    We exclude the Winograd schema datasets, WNLI and WSC, because they require a complete reformulation to get past the trivial most-frequent baseline \citep{kocijan-etal-2019-surprisingly}. The remaining 14 (Super)GLUE datasets measure performance on these tasks: 

    \begin{itemize}\itemsep0em 
    \item \textbf{Inference}: CB, MNLI, QNLI, RTE.
    \item \textbf{Linguistic acceptability}: CoLA.
    \item \textbf{Sentiment analysis}: SST-2.
    \item \textbf{Semantic similarity}: MRPC, QQP, STS-B.
    \item \textbf{Word sense disambiguation}: WiC.
    \item \textbf{Question answering}: BoolQ, COPA, MultiRC, ReCoRD.
    \end{itemize}

    
    \subsubsection{HANS}
    Deep learning systems are (by design) prone to finding spurious correlations in the training data.
    These heuristics can often be successfully employed for the evaluation data, as well -- thus, one has to be careful when implying that a higher score on a benchmark shows a deeper understanding of the tested model.
    \newcite{mccoy-etal-2019-right} tried to evaluate to what extent language models rely on spurious heuristics to solve NLI tasks. They identified a set of fallible syntactic heuristics and designed a test set where these `shortcuts' should fail -- Heuristic Analysis for NLI Systems (HANS). We adopt their approach and test models that have been fine-tuned on MNLI.

    \subsection{Edge probing}
    GLUE tasks measure the ability of a LM to be fine-tuned on a sentence-level NLU problem. To get a more comprehensive picture of LM performance, one can also \textit{probe} the word-level contextualized representations, measuring how much syntactic or semantic information can be extracted.

    \newcite{tenney2018what} devised a simple approach of probing for a diverse set of linguistic phenomena called \textit{edge probing}. They reformulate traditional NLP tasks as \textit{span classification}: part-of-speech tagging can be viewed as classification of word-spans and semantic role labeling becomes a classification of pairs of spans: predicate-span and argument-span. In the following, we will probe our models for five basic tasks: part-of-speech tagging (POS), dependency parsing (DP), semantic role labeling (SRL), named-entity recognition (NER) and coreference resolution (CR). Note that the model only learns to \textit{classify} each span provided to the model as gold data. This substantially simplifies some of the tasks, for example SRL. Please refer to \cref{sec:edge-probing-details} for the implementation details of edge probing.

    \subsection{BLiMP}
    \label{sec:blimp}
    
    One disadvantage of the aforementioned evaluation metrics is that the results are skewed by the second-stage supervised training, which makes it problematic to disentangle the prior knowledge of a language model from the acquired knowledge \cite{10.1162/coli_a_00422}. In contrast, the Benchmark of Linguistic Minimal Pairs \citep[BLiMP;][]{warstadt-etal-2020-blimp-benchmark} attempts to measure the linguistic knowledge of a language model in a zero-shot manner -- without any additional training. The dataset consists of 67\,000 sentence pairs; each pair differs minimally on the surface level, but only one of the sentences is grammatically valid. We can use the intrinsic ability of language models to assign a probability to every sentence and test how often a language model assigns a higher probability to the correct sentence. \cref{sec:blimp-details} gives more details about ranking the likelihood of sentences according to the raw output of a masked language model.
    
    \section{Experiments}

    We conduct a number of experiments in this section. First, we compare different training hyperparameters and model configurations described in \cref{sec:architecture}. Then, using the overall best training setting, we make a comparative study of training objectives (\cref{sec:objectives}). Finally, we investigate the sampling efficiency of our proposed language model and we compare BNC with a Wikipedia \& BookCorpus subset of the same size. These results can then be used as a baseline performance of BNC-BERT in future studies.
    
    The central model used in the experiments is a \textit{base}-sized Transformer -- 12 encoder layers with hidden size 768 and 12 attention heads (more details in \cref{sec:hyperparameters}). All reported models utilize the same cased WordPiece tokenizer \citep{https://doi.org/10.48550/arxiv.1609.08144} with a vocabulary size of $2^{14} = 16\,384$ trained with the BNC dataset (\cref{sec:tokenizer}). This goes against the trend of increasing the subword vocabulary in recent work,\footnote{BERT \citep{devlin-etal-2019-bert} uses 28\,996 tokens, RoBERTa \citep{DBLP:journals/corr/abs-1907-11692} 50\,265 and in 2021, DeBERTa \citep{he2021deberta} used a vocabulary of 128\,100 subwords.} but a larger vocabulary size would lead to a lot of infrequent tokens within our limited corpus -- we roughly follow \newcite{gowda-may-2020-finding} and \textit{`\dots use the largest possible BPE vocabulary such that at least 95\% of classes have 100 or more examples in training.'}
    
    
Since our aim is to train models comparable to $\textrm{BERT}_\textit{base}$, we train for the same amount of sampled tokens. \citet{devlin-etal-2019-bert} trained on 1M batches of 128K tokens, we use 31\,250 training steps with batch size of 4M tokens to parallelize and accelerate the process. Also, similarly to \newcite{devlin-etal-2019-bert}, we use sequence length of 128 tokens in the first 90\% of training and a larger sequence length of 512 only in the last 10\% of steps. We deliberately do not compare against more recent models, which are trained for much longer to achieve slightly better performance: RoBERTa is trained on $16\times$ more training samples, for example.\footnote{500K steps with 8\,192 segments of length 512, according to \cite{he2021deberta}.}
    
  
\subsection{Comparison of model architectures and training settings}
    
    In order to establish a strong baseline, we evaluate the proposed changes from \cref{sec:architecture} and other training configurations. We present the results in \cref{tab:comparison}, where we compare the final model with all changes applied and models with one of those modifications removed. These training choices turned out to be the most important:
    \begin{itemize}\itemsep0em
        \item Both the post-norm and pre-norm transformer variants perform substantially worse than the NormFormer-like layer normalization \citep{shleifer2022normformer}. Both of them also lead to less stable and only slightly faster training.
        \item Absolute positional embeddings seem to be less adaptable for fine-tuning but perform better on language modeling itself, as can be seen on the BLiMP results. We hypothesize that this is caused by more accurate estimation of probabilities of the first few words in a sentence. The simpler absolute embeddings also lead to the greatest reduction of training time. We choose the slower relative positional embeddings despite this fact to increase the performance on (Super)GLUE tasks.
        \item We observe that setting the weight decay correctly is crucial for masked language modeling. The default weight decay value found in \newcite{devlin-etal-2019-bert}, $0.01$, performs substantially worse on all tested tasks. We use a higher decay value of $0.1$ to boost performance, this value is most likely strongly correlated with the corpus size we use here. This suggests that previous findings of inferior performance of LMs pre-trained on small corpora might be caused by insufficient hyperparameter search.
        \item As expected, the AdamW optimizer \citep{loshchilov2018decoupled} behaves poorly in our highly parallel training regime. Our study successfully replicates the reported performance of the LAMB optimizer \citep{You2020Large}, which we thus use in all other experiments. 
    \end{itemize}

{\renewcommand{\arraystretch}{1.2}
\begin{table}[t]
\resizebox{\columnwidth}{!}{%
\begin{tabular}{@{}lrrrr@{}}
\toprule
\multirow{2}{*}{\textbf{Model}} & \multirow{2}{*}{\textbf{MNLI}} & \textbf{Edge} & \multirow{2}{*}{\textbf{BLiMP}} & \textbf{Training} \\ 
 & & \textbf{probing} & &  \textbf{time} \\ \midrule
LTG-BERT               & \textbf{85.1}$^{\pm0.2}$ & \textbf{95.3}$^{\pm0.1}$ & 83.4 & 8h 13min \\ \midrule
w/\hphantom{o} post-norm (0.005) & $-0.5^{\pm0.2}$ & $-0.6^{\pm0.1}$ & $-0.1$ & \hphantom{0h }$-$22min\\
w/\hphantom{o} pre-norm (0.005)  & $-1.3^{\pm0.1}$ & $-0.2^{\pm0.1}$ & $-0.9$ & \hphantom{0h }$-$35min \\
w/\hphantom{o} GELU activation      & $-0.3^{\pm0.3}$ & \hphantom{$-$}\textbf{0.0}$^{\pm0.1}$ & $-0.1$ & \hphantom{1h1 }$-$6min \\
w/\hphantom{o} absolute pos. emb.   & $-1.1^{\pm0.2}$   &   $-$\textbf{0.1}$^{\pm0.1}$    &       $+$\textbf{0.6}  & $-$\textbf{2h 16min} \\
w/o FF init. scaling     &  $-0.3^{\pm0.2}$     &     $-$\textbf{0.1}$^{\pm0.1}$     &   $+0.1$ & \hphantom{$-0h 0$}0min \\
w/\hphantom{o} learnt FF biases        &  $-0.3^{\pm0.2}$     &     \hphantom{$-$}\textbf{0.0}$^{\pm0.1}$     &  $-0.1$ & \hphantom{0h 0}$+$9min  \\
w/\hphantom{o} 0.01 WD (0.005)   &  $-1.4^{\pm0.1}$     &     $-0.2^{\pm0.1}$       &     $-0.7$ & \hphantom{0h 0 }$-$1min  \\
w/\hphantom{o} linear schedule      &  $-0.5^{\pm0.2}$     &     \hphantom{$-$}\textbf{0.0}$^{\pm0.1}$       &     $-0.2$ & \hphantom{$-$0h 0}0min  \\
w/\hphantom{o} AdamW (0.001)     & $-0.9^{\pm0.2}$ & $-0.2^{\pm0.1}$ & $-0.5$    & \hphantom{0h }$-$11min         \\ \bottomrule
\end{tabular}%
}
\caption{Comparative study of different architectural and training settings. The first row shows the performance of the final model with all improvements applied and the following rows give the relative changes in performance when one of the changes is not applied -- for example, the second row tests swapping the NormFormer-like normalization with the `post-norm' normalization. Some runs diverged with the default learning rate of $0.01$ and had to be run again with a lower value (denoted in parentheses). `WD' stands weight decay and `FF' is an abbreviation for the feed-forward modules. We report the mean and standard deviation statistics across five runs, if applicable, and boldface all run within 1 standard deviation from the best result.}
\label{tab:comparison}
\end{table}
}

{\renewcommand{\arraystretch}{1.15}
\begin{table*}[!th]
    \resizebox{\textwidth}{!}{%
    \begin{tabular}{@{}l@{\hspace{1.0em}}ccccc@{\hspace{2.5em}}l@{\hspace{2.5em}}c@{\hspace{2.0em}}c@{}}
    \toprule
    \multicolumn{1}{@{}l}{\multirow{2}{*}{\textbf{Model (variant)}}} & \multicolumn{5}{@{}c@{}}{\hspace{-3em}\crulefill~~\textbf{GLUE}~~\crulefill\hspace{2em}}                                       & \multirow{2}{*}{\textbf{HANS}} & \textbf{Edge} & \multirow{2}{*}{\textbf{BLiMP}} \\
    \multicolumn{1}{l}{}                                & \footnotesize{\textbf{MNLI}} & \footnotesize{\textbf{MRPC}} & \footnotesize{\textbf{QNLI}} & \footnotesize{\textbf{SST-2}} & \footnotesize{\textbf{Average}} &                                &      \textbf{probing}                                  &                                 \\ \midrule
    \multicolumn{9}{@{}c@{}}{\raisebox{0.8ex}{\footnotesize{Wikipedia + BookCorpus \citep[3000M words;][]{devlin-etal-2019-bert}}}} \\
    \multicolumn{1}{@{}l}{$\textrm{BERT}_\textit{base, cased} {}^\dagger$}                                                       & 84.4\hphantom{$^{\pm0.1}$}          & 86.7\hphantom{$^{\pm0.1}$}        & 88.4\hphantom{$^{\pm0.1}$}          & 92.7\hphantom{$^{\pm0.1}$}         &      88.1\hphantom{$^{\pm0.1}$}           &               \textbf{69.0}\hphantom{$^{\pm0.1}$}                 & 93.9\hphantom{$^{\pm0.1}$}                                   & \textbf{84.2}                            \\
        \multicolumn{1}{@{}l}{$\textrm{BERT}_\textit{base, cased}\,\,\textrm{(our eval.)}$}                                                & 83.6$^{\pm0.2}$ &	84.6$^{\pm0.5}$	& 90.8$^{\pm0.1}$ &	91.9$^{\pm0.4}$ & 87.8$^{\pm0.3}$ & 61.8$^{\pm1.5}$ &	93.8$^{\pm0.2}$	& \textbf{84.2}                            \\ \midrule
    \multicolumn{9}{@{}c@{}}{\raisebox{0.8ex}{\footnotesize{Wikipedia + BookCorpus (100M words)}}} \\

 LTG-BERT (subword masking)         & 84.2$^{\pm0.1}$ & 84.3$^{\pm0.7}$ & 90.8$^{\pm0.3}$ & 92.1$^{\pm0.5}$ &  87.8$^{\pm0.5}$ & 62.5$^{\pm1.7}$ & \textbf{95.3}$^{\pm0.1}$ & 82.0 \\ \midrule
    
    \multicolumn{9}{@{}c@{}}{\raisebox{0.8ex}{\footnotesize{British National Corpus (100M words)}}} \\

    LTG-BERT (subword masking)         & \textbf{85.1}$^{\pm0.2}$ & 85.0$^{\pm0.9}$ & 90.0$^{\pm0.3}$ & \textbf{92.7}$^{\pm0.4}$ & 88.2$^{\pm0.5}$ & 64.4$^{\pm1.3}$ & \textbf{95.3}$^{\pm0.1}$ & 83.4 \\
    LTG-BERT (whole-word masking)    & 84.9$^{\pm0.2}$ & 85.5$^{\pm0.9}$ & 90.6$^{\pm0.3}$ & \textbf{92.7}$^{\pm0.2}$ & 88.4$^{\pm0.5}$ & 63.7$^{\pm0.8}$ & \textbf{95.3}$^{\pm0.1}$ & 80.1 \\
    LTG-BERT (span masking)       & \textbf{85.1}$^{\pm0.2}$ & \textbf{87.5}$^{\pm0.9}$ & \textbf{91.5}$^{\pm0.2}$ & \textbf{92.8}$^{\pm0.5}$ & \textbf{89.2}$^{\pm0.5}$ & 65.6$^{\pm0.5}$ & \textbf{95.2}$^{\pm0.1}$ & \textbf{84.2} \\ \midrule
LTG-BERT (subword + document NSP) & \textbf{85.2}$^{\pm0.3}$ & 86.5$^{\pm0.8}$ & 90.3$^{\pm0.2}$ & 92.2$^{\pm0.4}$ & 88.6$^{\pm0.5}$ & 60.5$^{\pm1.2}$ & \textbf{95.3}$^{\pm0.1}$ & 83.3 \\
LTG-BERT (subword + order NSP)  & 84.7$^{\pm0.1}$ & 85.9$^{\pm0.6}$ & 90.4$^{\pm0.2}$ & 92.1$^{\pm0.2}$ &  88.3$^{\pm0.4}$ & 64.2$^{\pm1.9}$ & 95.1$^{\pm0.1}$ & 82.2 \\ \midrule
LTG-BERT (subword + $2\times$ steps) & \textbf{85.2}$^{\pm0.2}$ & 86.5$^{\pm0.8}$ & 90.3$^{\pm0.3}$ & \textbf{92.3}$^{\pm0.6}$ & 88.6$^{\pm0.5}$ & 65.3$^{\pm1.1}$ & \textbf{95.3}$^{\pm0.1}$ & 83.5 \\
LTG-BERT (subword + $\nicefrac{1}{2}\times$ steps)   & 84.4$^{\pm0.3}$ & 86.3$^{\pm1.1}$ & 90.4$^{\pm0.2}$ & \textbf{92.8}$^{\pm0.4}$ & 88.5$^{\pm0.6}$ & 62.4$^{\pm0.8}$ & \textbf{95.2}$^{\pm0.1}$ & 83.5 \\
LTG-BERT (subword + $\nicefrac{1}{4}\times$ steps)  & 83.8$^{\pm0.2}$ & 85.3$^{\pm0.8}$ & 89.1$^{\pm0.2}$ & 91.7$^{\pm0.4}$ & 87.5$^{\pm0.5}$ & 58.6$^{\pm1.3}$ & 95.0$^{\pm0.1}$ & 83.2 \\ \midrule
\multicolumn{1}{@{}l}{Random initialization}                        & 59.5$^{\pm0.5}$ &	68.5$^{\pm1.4}$ &	63.8$^{\pm0.2}$ & 82.2$^{\pm0.7}$ & 68.5$^{\pm0.8}$ &  49.7$^{\pm0.3}$ &	73.1$^{\pm0.4}$	& 50.0                            \\ \bottomrule
    \end{tabular}%
    }
    \caption{Summary of the experimental results. We show the results on the 4 GLUE tasks with known development results from \newcite{devlin-etal-2019-bert} and their average; then the accuracy on HANS, the average of all 5 edge probing tasks and 67 BLiMP tasks. $^\dagger$The $\textrm{BERT}_\textit{base, cased}$ results are shown primarily for reference, they come from these sources: partial development GLUE scores from \newcite{devlin-etal-2019-bert}, edge probing from \newcite{tenney2018what}, HANS from \newcite{bhargava-etal-2021-generalization} and BLiMP from \newcite{salazar-etal-2020-masked}. We also add the $\textrm{BERT}_\textit{base, cased}$ results from our evaluation scripts for more fair and accurate comparison. We present the mean and standard deviation statistics over 5 evaluation runs and boldface all run within 1 standard deviation from the best result. The detailed results can be found in \cref{sec:detailed-results}.}
    \label{tab:results}
\end{table*}
}    

    \noindent
    The other changes bring more marginal gains -- all three tested modifications of the feed-forward layers work slightly better: 1) using GEGLU activation function instead of GELU, 2) initializing the feed-forward layers with incrementally lower weight norms, and 3) not using any bias parameters in these layers. The last tested change shows that cosine learning rate decay \citep{https://doi.org/10.48550/arxiv.2112.11446} performs better than the standard linear weight decay.

    \subsection{Training objective comparison}

    \paragraph{Masked language modeling.}
    First of all, we compare the three masking methods described in \cref{sec:mlm}: subword, whole-word and span masking. The summary of the results is given in \cref{tab:results}, more detailed evaluation in \cref{sec:detailed-results}. Overall, the span-based masking performs marginally better than the other methods -- it shows a clear improvement on (Super)GLUE benchmarks over the simple subword masking, it generalizes the best according to the HANS score and it even matches the performance of $\textrm{BERT}_\textit{base}$ on the averaged BLiMP accuracy. All methods perform equally well on edge probing. Whole-word masking lacks on the BLiMP benchmark because the model is not expecting partially masked words that can occur in the evaluation (\cref{sec:blimp}). The original subword masking strategy is still a competitive baseline and it might be preferred in practice due to its simple implementation.  
    
    \paragraph{Next-sentence prediction.} Next, we experiment with combining an NSP task and simple subword masking. We hypothesize that a second training objective might extract more information from the limited BNC corpus, which would help with the downstream performance -- an opposite conclusion than \newcite{DBLP:journals/corr/abs-1907-11692}. However, our hypothesis turns out to be wrong, according to the results in \cref{tab:results}. The experiments agree with the design of latest masked language models -- next sentence prediction is an unnecessary training objective, at least for the tasks evaluated in this paper. It does not lead to substantially improved sentence representations even in a limited data regime. We can also see that the well-motivated order discrimination \citep{Lan2020ALBERT}, proposed to solve the issues of document discrimination, actually leads to an overall worse performance. Hence we cannot recommend to complicate pre-training with a second training objective.

    \subsection{Sampling efficiency}
    \label{sec:sample-efficiency}
    
    An important aspect of efficient language models is the number of training steps they require to reach a sufficient performance. So far, we have limited the size of the training corpus but kept the number of steps constant, set according to \newcite{devlin-etal-2019-bert}. The results in \cref{tab:results} suggest that increasing the steps two times does not lead to a noticeably better performance with BNC. Even more so, training for half the time turns out to be enough to get comparable performance. Yet, decreasing the training steps further starts to degrade the downstream results too much, as evidenced by the scores obtained with $\nicefrac{1}{4}$ of the default steps.
    
    These results highlight the sampling inefficiency of current self-supervised language modeling methods, as even with $\nicefrac{1}{4}$ steps, every token in BNC is seen about roughly 250 times during training.\footnote{This value can be calculated from \cref{tab:hyperparams}: these models are trained for 7\,812 steps with 4\,194\,304 tokens per batch. \cref{tab:bnc} shows that there are 131\,392\,103 subwords in the BNC train split.} We hope that a future work in this field will be able to learn from a smaller number of samples.
    
    \subsection{100 million subset of Wikipedia \& BookCorpus}
    \label{sec:wiki-subset-eval}
    
    Our last experiment evaluates how much does the careful curation of BNC help the downstream performance. To keep the comparability to BERT, we choose to pre-train on a random subset of Wikipedia and BookCorpus (with equal size to BNC, sampled document-wise); this corpus is constructed according to \cref{sec:wiki}. Note that BNC is a corpus of British English compiled in 1990s so some evaluation tasks can be skewed against it -- for example QNLI, which is based on texts from Wikipedia. \cref{tab:results} shows that a high-quality data source is not necessarily needed to learn from 100M words but better quality leads to a noticeable difference in downstream performance.

\section{Conclusion}
In this paper, we evaluated how data-efficient masked language models can be. In particular, we trained a variety of models with different training objectives on the same training data: British National Corpus. Although small by modern standards (100M tokens), it is well balanced and carefully crafted to represent British English of the 20\textsuperscript{th} century. On a variety of benchmarks, our models perform better than BERT\textsubscript{\textit{base}} trained on a much larger corpus. We believe that this limited data regime is beneficial for the development of efficient and reliable language models. Our finding also suggests that 100 million word tokens is enough to learn basic linguistic skills by current language modeling techniques, given that the data is carefully selected and balanced. To conclude, huge amounts of training data are not always necessary -- we should focus on more efficient training settings instead.

We showed that the next sentence prediction objective does not improve BERT-like models, confirming the findings in \newcite{DBLP:journals/corr/abs-1907-11692}. In addition, the standard subword masking from \newcite{devlin-etal-2019-bert} is consistently outperformed by the span masking method and the linguistic performance can be substantially increased by utilizing better neural architectures and training configurations. We release the code for training and using BERT-like models with the optimal architectural choices (according to our experiments) under the name LTG-BERT.\footnote{\url{https://github.com/ltgoslo/ltg-bert}}
    
The presented results serve primarily as the foundation for future research on efficient language modeling. We hope our work shows the value of careful curation of representative corpora and will spark more interest in this area, where BNC can serve as an undemanding, replicable and openly-available training corpus.

\section{Limitations}

First of all, our work only considers language modeling of English and does not provide results on any other language -- even though we hope that our conclusions could be useful for low-resource languages. Secondly, even though we found out that it is possible to train a competent language model with a small corpus, the training process still requires a similar amount of computational resources to models trained with larger corpora, as noted in \cref{sec:sample-efficiency}. Finally, we evaluate mainly the linguistic knowledge of language models (\cref{sec:metrics}), our conclusions might not apply for their general knowledge.

\section*{Acknowledgement}
The computations were performed on resources provided by Sigma2 -- the National Infrastructure for High Performance Computing and Data Storage in Norway.

    
    \bibliography{anthology,custom}
    \bibliographystyle{acl_natbib}
    
    \clearpage

    \appendix

{\renewcommand{\arraystretch}{1.1}
\begin{table*}[!ht]
\resizebox{\textwidth}{!}{%
\begin{tabular}{@{}lrrrrrrrrrrrrrr@{}}
\toprule
\textbf{Task}          & \textbf{BoolQ} & \textbf{CB} & \textbf{CoLA} & \textbf{COPA} & \textbf{MNLI} & \textbf{MRPC} & \textbf{MultiRC} & \textbf{QNLI} & \textbf{QQP} & \textbf{ReCoRD} & \textbf{RTE} & \textbf{SST-2} & \textbf{STS-B} & \textbf{WiC} \\ \midrule
Train size             & 9\,427           & 250         & 8\,551          & 800           & 392\,702        & 3\,668          & 27\,243            & 104\,743        & 363\,846       & 1\,179\,400         & 2\,490         & 67\,349         & 5\,749          & 5\,428         \\
Validation size        & 3\,270           & 56          & 1\,043          & 100           & 9\,815     & 408           & 4\,848             & 5\,463          & 40\,430        & 113\,236          & 277          & 872           & 1\,500          & 638          \\
$\ge$ 512 subwords & 0.37\%     & 0\%     & 0\%        & 0\%        & 0\%        & 0\%        & 27.68\%         & 0.02\%        & 0\%       &  0.30\%        & 0\%      & 0\%        & 0\%        & 0\%       \\ \bottomrule
\end{tabular}%
}
\caption{The train and validation sizes of GLUE and SuperGLUE tasks (omitting WNLI and WSC). Note that we list the numbers of examples in the (Super)GLUE formulation of these tasks, which may differ from the actual number of examples -- for example in case of multiple-choice questions. Some tasks do not offer a reliable amount of training data and some tasks contain a large number of examples longer than the length limit of our language models.}
\label{tab:glue}
\end{table*}
}

\renewcommand{\arraystretch}{1.5}
\begin{table*}[ht!]
\resizebox{\textwidth}{!}{%
\begin{tabular}{@{}lrrrrrrrrrrrrr@{}}
\toprule
\multirow{2}{*}{\textbf{Task}} & \multicolumn{12}{@{}c@{}}{\crulefill\crulefill\crulefill\crulefill\crulefill~~~\textbf{Layer}~~~\crulefill\crulefill\crulefill\crulefill\crulefill}                                                                                                                                                                                                                                                                                                                 & \textbf{Regression} \\
                               & \hspace{2em}\textbf{1}                & \hspace{2em}\textbf{2}                 & \hspace{2em}\textbf{3}                 & \hspace{2em}\textbf{4}                & \hspace{2em}\textbf{5}                & \hspace{2em}\textbf{6}                 & \hspace{2em}\textbf{7}                 & \hspace{2em}\textbf{8}                 & \hspace{2em}\textbf{9}                 & \hspace{2em}\textbf{10}               & \hspace{2em}\textbf{11}                & \hspace{2em}\textbf{12}              &                                \textbf{slope} \\ \midrule
POS   & \cellcolor{blue!20.00}27.49 & \cellcolor{blue!7.06}12.55  & \cellcolor{blue!2.72}7.54   & \cellcolor{blue!0.88}5.42   & \cellcolor{blue!1.19}5.78   & \cellcolor{blue!0.38}4.83   & \cellcolor{blue!0.63}5.12   & \cellcolor{blue!1.03}5.59  & \cellcolor{blue!1.41}6.02 & \cellcolor{blue!0.89}5.43 & \cellcolor{blue!0.00}4.40 & \cellcolor{blue!4.71}9.84   & -0.98 \\
DP    & \cellcolor{blue!20.00}14.65 & \cellcolor{blue!12.81}10.78 & \cellcolor{blue!13.21}10.99 & \cellcolor{blue!16.31}12.66 & \cellcolor{blue!11.22}9.92  & \cellcolor{blue!6.65}7.47   & \cellcolor{blue!7.63}8.00   & \cellcolor{blue!4.00}6.04  & \cellcolor{blue!3.29}5.66 & \cellcolor{blue!1.28}4.58 & \cellcolor{blue!0.00}3.89 & \cellcolor{blue!2.71}5.35   & -0.89 \\
SRL   & \cellcolor{blue!20.00}19.38 & \cellcolor{blue!13.19}13.70 & \cellcolor{blue!8.52}9.80   & \cellcolor{blue!8.23}9.56   & \cellcolor{blue!6.88}8.44   & \cellcolor{blue!5.21}7.04   & \cellcolor{blue!5.15}6.99   & \cellcolor{blue!3.49}5.61  & \cellcolor{blue!2.59}4.85 & \cellcolor{blue!1.36}3.83 & \cellcolor{blue!0.00}2.70 & \cellcolor{blue!6.49}8.11   & -1.04 \\
NER   & \cellcolor{blue!20.00}18.16 & \cellcolor{blue!6.43}9.12   & \cellcolor{blue!2.63}6.58   & \cellcolor{blue!0.00}4.83   & \cellcolor{blue!3.05}6.86   & \cellcolor{blue!2.89}6.75   & \cellcolor{blue!3.06}6.87   & \cellcolor{blue!2.18}6.28  & \cellcolor{blue!2.69}6.62 & \cellcolor{blue!0.16}4.93 & \cellcolor{blue!1.56}5.87 & \cellcolor{blue!18.45}17.13 & -0.16 \\
COREF & \cellcolor{blue!7.92}7.24   & \cellcolor{blue!11.97}9.12  & \cellcolor{blue!9.08}7.78   & \cellcolor{blue!14.95}10.50 & \cellcolor{blue!17.93}11.89 & \cellcolor{blue!20.00}12.85 & \cellcolor{blue!18.94}12.35 & \cellcolor{blue!11.63}8.96 & \cellcolor{blue!3.53}5.20 & \cellcolor{blue!1.54}4.27 & \cellcolor{blue!0.00}3.56 & \cellcolor{blue!5.88}6.29   & -0.42 \\ \bottomrule
\end{tabular}%
}
\caption{The per-layer contributions to different edge probing tasks, taken from the layer-wise convex weights $\gamma_k$ (rendered in percent). To summarize the individual scores, we fit a linear regression line and show its slope in the last column. A negative slope implies stronger representation in the lower layers and vice versa.}
\label{tab:layers}
\end{table*}
\renewcommand{\arraystretch}{1}

\section{BNC samples}
\label{sec:bnc-samples}

We follow the description of the Markdown conversion of BNC from \cref{sec:preprocessing} and show samples of the resulting raw Markdown text to illustrate this process, highlighting some of the formatting information captured by our format. A sample of a spoken document is given in \cref{list:spoken} and a sample of a written article is shown below, in \cref{list:written}.

\section{Evaluation metrics -- implementation details}
\label{sec:evaluation-details}

\subsection{(Super)GLUE}
\label{sec:superglue-details}
Fine-tuning of the GLUE and SuperGLUE tasks follows a straightforward framework: the segments are tokenized, concatenated -- starting with a special \texttt{[CLS]} token and with a \texttt{[SEP]} token put in between the segments -- and input to a pre-trained language model. Subsequently,  
    the contextualized representation of the special 
    \texttt{[CLS]} token  
    is fed into an MLP classifier. The pre-trained weights are further fine-tuned together with the classifier weights.
    
    We do not employ any additional training tricks that were used in the previous works to seemingly increase the performance of their large language models -- e.g. further `pre-training' on MNLI, multi-task learning, ensembling, extensive hyperparameter search, selecting the best random seeds, reformulating the tasks or complex regularization techniques such as SMART \citep{jiang-etal-2020-smart}.
    
\subsection{Edge probing}
\label{sec:edge-probing-details}
We follow the description of edge probing in the original paper by \newcite{tenney2018what}. First of all, subword representations $s_{i,k}$ are extracted from a frozen LM, for all positions $i$ and layers $k$. These are downsampled to a dimensionality of 256 by a linear transformation. To get a vector representation $h_t$ for the $t^\textrm{th}$ span, we apply two pooling operations on the subword-token representations $s_{t,k}$. First, we pool the vectors at all layers $k$ by taking a learnt convex combination $\hat{s}_t = \sum_{k=1}^{12}{\gamma_k s_{t,k}}$, where $\gamma_k \in \mathbb{R}$. Next, since one span can be split into multiple subwords, we employ an attention pooling operator to get the span-level embeddings: $h_t = \sum_{i \in \mathcal{I}_t}{\textrm{att}(\hat{s}_t; \theta)\hat{s}_t}$, where $\mathcal{I}_t$ are the subwords indices of the $t^\textrm{th}$ span. Finally, the pooled vectors $h_t$ are fed into a multi-layer perceptron (MLP) and classified. If a task requires a pair of span representations (DP, SRL and CR), then these are pooled with two separate attention operators and concatenated before being passed to the MLP classifier.

\begin{figure}[t!]
        \centering
        \includegraphics[width=\columnwidth]{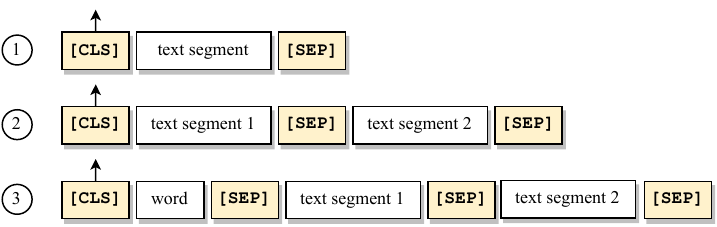}
        \caption{Three variations of (Super)GLUE input: 1) single-sentence tasks SST-2 and CoLA; 2) classification of a pair of text segments: BoolQ, CB, COPA, MNLI, MRPC, QNLI, QQP, STS-B, RTE; and 3) WiC (in the figure), MultiRC, ReCoRD.}
        \label{fig:glue}
    \end{figure}

\subsection{BLiMP}
\label{sec:blimp-details}
These models are trained to estimate $P(\bm{s}_t \vert \bm{s}_{<t})$ for sentence $\bm{s}$ and token $\bm{s}_t$ where $\bm{s}_{<t} = \left(\bm{s}_i \vert i < t\right)$; then the sentence log-probability is given by $\textrm{log}P(\bm{s})=\sum_{t=1}^{N}{\textrm{log}P(\bm{s}_t \vert \bm{s}_{<t})}$.
    
    The issue with masked language models is that they are not designed to calculate this property; they are trained to estimate $P(\bm{s}_t \vert \bm{s}_{\setminus t})$ -- the likelihood of a token $s_t$ given its bidirectional context $\bm{s}_{\setminus t} = \left(\bm{s}_i \vert i \neq t\right)$. We can however still use MLMs to infer a \textit{score} for each sentence where a higher \textit{score} corresponds to a more likely sentence. \newcite{wang-cho-2019-bert} defined \textit{pseudo-log-likelihood score} of a sentence $s$ with model $\theta$ as
    $$\textrm{PLL}(\bm{s})=\sum_{t=1}^{N}{\textrm{log}P(\bm{s}_t \vert \bm{s}_{\setminus t}; \theta)}.$$
    \newcite{salazar-etal-2020-masked} tested PLL and found that it produces accurate predictions on BLiMP. We adopt their approach and evaluate our models with PLL.

\section{Layer interpretation}

The definition of the fine-tuning scheme for edge probing makes it straightforward to rate the contribution of each Transformer layer to a particular task -- we can simply have a look at the layer-wise weights $\gamma_k$, see \cref{tab:layers}. To be more precise, if we define the $k$\textsuperscript{th} attention layer as $a_k$, the $k$\textsuperscript{th} feed-forward layer as $f\!f_k$ and layer normalization operator as LN, then the $k$\textsuperscript{th} layer $\ell_k$ of a post-norm Transformer \citep{NIPS2017_3f5ee243} computes the following function:
\begin{equation*} 
\begin{split}
\hat{a}_k(x) & = \textrm{LN}\!\left(x + a_k(x)\right) \\
\ell_k & = \textrm{LN}\bigl(\hat{a}_k(\ell_{k-1}) + f\!f_k(\hat{a}_k(\ell_{k-1}))\bigr) \\
\end{split}
\end{equation*}

It is unclear how to separate the contribution of each layer from the the previous layers here: $\ell_k$ contains both the previous scaled $\ell_{k-1}$ and its transformation from $a_k$ and $f\!f_k$. On the other hand, our NormFormer-like architecture (\cref{sec:architecture}) defines each layer $\ell_k$ as:
\begin{equation*} 
\begin{split}
\hat{a}_k(x) & = x + \textrm{LN}\!\left(a_k(x)\right) \\
\ell_k & = \hat{a}_k(\ell_{k-1}) + f\!f_k(\hat{a}_k(\ell_{k-1}))\bigr) \\
\end{split}
\end{equation*}

Then it is trivial to calculate the contribution of each layer as $s_k = \ell_k - \ell_{k-1}$. We use this $s_k$ entities to compute the learnt convex combination of all layers $\hat{s} = \sum_{k=1}^{12}{\gamma_k s_{k}}$. 

Interpreting $\gamma_k$ as the amount of `knowledge' of a particular task in layer $k$, we see that POS information is contained primarily in the lowest layers, followed by SRL and DP. On the other hand, NER and CR are represented more strongly in the higher layers, which confirms the related findings in the literature \citep{rogers-etal-2020-primer}.

\section{Fine-grained results}

\label{sec:detailed-results}
To ease the evaluation of any future language models trained on BNC, we provide detailed results of all evaluated models in the following tables: GLUE results are shown in \cref{tab:finegrained-glue}, edge probing performance is given in \cref{tab:edge-probing-finegrained} and the BLiMP accuracies in \cref{tab:blimp-detailed}.

\paragraph{(Super)GLUE.} In total, we fine-tune all models on these 14 (Super)GLUE datasets:  

\begin{itemize}\itemsep0em 
    \item \textbf{Boolean Questions} \citep[BoolQ;][]{clark-etal-2019-boolq}, a yes/no question answering dataset evaluated with accuracy.
    \item \textbf{The CommitmentBank} \citep[CB;][]{de_Marneffe_Simons_Tonhauser_2019}, evaluated with both accuracy and F\textsubscript{1}-score, where the multi-class F\textsubscript{1}
is computed as the unweighted average of the F\textsubscript{1} per class.
    \item \textbf{Corpus of Linguistic Acceptability} \citep[CoLA;][]{warstadt-etal-2019-neural} evaluated with the Matthews correlation coefficient \citep[MCC;][]{MATTHEWS1975442}.
    \item \textbf{Choice of Plausible Alternatives} \citep[COPA;][]{copa}, evaluated with accuracy.
    \item \textbf{The Multi-Genre Natural Language Inference Corpus} \citep[MNLI;][]{williams-etal-2018-broad}. Its development set consists of two parts: \textit{matched}, sampled from the same data source as the training set, and \textit{mismatched}, which is sampled from a different domain. Both parts are evaluated with accuracy.
    \item \textbf{The Microsoft Research Paraphrase Corpus} \citep[MRPC;][]{dolan-brockett-2005-automatically}, evaluated with both accuracy and F\textsubscript{1}-score.
    \item \textbf{Multi-Sentence Reading Comprehension} \citep[MultiRC;][]{khashabi-etal-2018-looking}, a multiple choice question answering dataset, evaluated with the exact match accuracy (EM) and F\textsubscript{1}-score (over all answer options).
    \item \textbf{Question-answering Natural Language Inference} (QNLI) constructed from the Stanford Question Answering Dataset \citep[SQuAD;][]{rajpurkar-etal-2016-squad}, evaluated with accuracy.
    \item \textbf{The Quora Question Pairs} (QQP),\footnote{\url{https://quoradata.quora.com/First-Quora-Dataset-Release-Question-Pairs}} evaluated with both accuracy and F\textsubscript{1}-score.
    \item \textbf{The Stanford Sentiment Treebank} \citep[SST-2;][]{socher-etal-2013-recursive}, evaluated with accuracy.
    \item \textbf{The Semantic Textual Similarity Benchmark} \citep[STS-B;][]{cer-etal-2017-semeval}, evaluate with Pearson and Spearman correlation coefficients.
    \item \textbf{Reading Comprehension with Commonsense Reasoning Dataset} \citep[ReCoRD;][]{https://doi.org/10.48550/arxiv.1810.12885}, a question answering dataset evaluated with the exact match accuracy (EM) and token-level F\textsubscript{1}-score (maximum over all correct mentions).
    \item \textbf{The Recognizing Textual Entailment datasets} \citep[RTE;][]{10.1007/11736790_9, rte2, giampiccolo-etal-2007-third, Bentivogli09thefifth}, evaluated with accuracy.
    \item \textbf{The Word-in-Context dataset} \citep[WiC;][]{pilehvar-camacho-collados-2019-wic}, evaluated simply with accuracy.
\end{itemize}

\paragraph{Edge probing.} We report the results on part-of-speech tagging (POS), semantic role labeling (SRL), named entity recognition (NER) and coreference resolution (CR) using the annotations from the English part of OntoNotes 5.0 \citep{https://doi.org/10.35111/xmhb-2b84}. In addition, to further measure the syntactic abilities, we test the dependency parsing (DP) accuracy on the English Web Treebank v2.9 dataset from the Universal Dependencies \citep{silveira14gold}.\footnote{Available online at \url{https://github.com/UniversalDependencies/UD_English-EWT}.} These choices follow the original work by \newcite{tenney2018what}, but we do not evaluate on constituency parsing, because the results suffered from large variation. Instead, we test the syntactic knowledge with DP, which turned out to be more reliable as its variation is negligible (\cref{tab:edge-probing-finegrained}).

\paragraph{BLiMP.} The Benchmark of Linguistic Minimal Pairs for English \citep{warstadt-etal-2020-blimp-benchmark} consists of 67 tasks. Each focuses on a specific linguistic feature, which is tested with 1\,000 automatically generated sentence pairs. \newcite{warstadt-etal-2020-blimp-benchmark} clusters these tasks into the following subgroups:
\begin{itemize}\itemsep0em
\item \textbf{Anaphor agreement} tests whether the reflexive pronouns agree with their antecedents.
\item \textbf{Argument structure} -- do verbs appear with the correct types of arguments?
\item \textbf{Binding} evaluates the correctness of structural relationship between a pronoun and its antecedent.
\item \textbf{Control/raising} tests syntactic and semantic differences between predicates
that embed an infinitival verb predicate.
\item \textbf{Determiner-noun agreement} checks number
agreement between determiners the associated noun.
\item \textbf{Ellipsis} -- can we omit an expression from a sentence?
\item \textbf{Filler-gap} tests dependencies created by phrasal movement.
\item \textbf{Irregular forms} checks the correctness of irregular morphology on
English past participles.
\item \textbf{Island effects} -- correctness of a possible gap in a filler-gap dependency.
\item \textbf{NPI licensing} -- are the negative polarity items used correctly (e.g. in negation)?
\item \textbf{Quantifiers} tests the usage of quantifiers.
\item \textbf{Subject-verb agreement} checks the number agreement between present tense verbs and subjects.
\end{itemize}

\section{Hyperparameters}
\label{sec:hyperparameters}

All hyperparameters used to pre-trained and fine-tune our models are listed below: pre-training hyperparameters in \cref{tab:hyperparams}, the GLUE and SuperGLUE fine-tuning hyperparameters in \cref{tab:glue-hyperparams} and the edge probing hyperparameters in \cref{tab:ep-hyperparams}. BLiMP does not require any special hyperparameters, it need only out-of-the-box predictions of a pre-trained language model. Note that we will also release the full PyTorch \citep{NEURIPS2019_9015} source code, tokenizer and the pre-trained language models in the camera-ready version. Additionaly, we will also provide all necessary wrappers for a simple use of our models with the \texttt{transformers} library \citep{wolf-etal-2020-transformers}.

The training was performed on 128 AMD MI250X GPUs (distributed over 16 compute nodes) and took approximately 8 hours per model in a mixed precision mode. In total, our models consist of 98M parameters; a slightly lower value than BERT's 110M parameters due to the smaller vocabulary size.

\section{Wikipedia + BookCorpus dataset replication}
\label{sec:wiki}

The information about the exact Wikipedia dump used for training BERT is unknown and the BookCorpus dataset \citep{7410368} is no longer available. On top of that, the preprocessing choices are also not known. Our 100M Wikipedia + BookCorpus dataset is thus different from the original BERT pre-training corpus.

We downloaded a fresh English Wikipedia dump from {\footnotesize \url{https://dumps.wikimedia.org/enwiki/20220801/enwiki-20220801-pages-articles-multistream.xml.bz2}}, extracted the raw text with WikiExtractor \citep{Wikiextractor2015} and segmented each article into sentences with \texttt{spaCy}.\footnote{\url{https://spacy.io/}}

A replicated version of BookCorpus was obtained from {\footnotesize \url{https://the-eye.eu/public/AI/pile_preliminary_components/books1.tar.gz}} and every book was also segmented with \texttt{spaCy}.

After that, the random 100M subset was created by sampling random documents from the full Wikipedia + BookCorpus dataset until the subset contained as many characters as BNC.

\section{Tokenizer definition}
\label{sec:tokenizer}

We use the HuggingFace's \texttt{tokenizers} library,\footnote{\url{https://huggingface.co/tokenizers/}} to define and train a subword tokenizer on BNC (training split).\footnote{We share the full definition of the tokenizer in \url{https://github.com/ltgoslo/ltg-bert}.}

Following the suggestion of \newcite{gowda-may-2020-finding}, we set the vocabulary size so that at least 95\% of tokens appear more than 100 times. In our case, with the size of $2^{14} = 16\,384$, 95\% of tokens appear more than 166 times in the training split. Their finding comes from the realm of neural machine translation, we have not evaluated how it aligns with language modeling. Nevertheless, we believe that a comparative study of different tokenizer settings makes an interesting future work; we suspect that the effects will be more pronounced with BNC, due to its limited size.

\begin{table*}
\centering
\resizebox{\textwidth}{!}{%
\begin{tabular}{@{}llr@{\hspace{2.0em}}ccc@{\hspace{2.0em}}cc@{\hspace{2.0em}}ccc@{}}
\toprule
\multirow{2}{*}{\textbf{Task}} & \multirow{2}{*}{\textbf{Metric}} & \multirow{2}{*}{\hspace{-2.5em}\textbf{BERT (100M subset)}}        & \multicolumn{3}{c}{\hspace{-2.0em}\crulefill~~\textbf{MLM}~~\crulefill}                                                     & \multicolumn{2}{c}{\hspace{-2.0em}\crulefill~~\textbf{NSP}~~\crulefill}                        & \multicolumn{3}{c}{\crulefill~~\textbf{Training steps}~~\crulefill}           \\
                               &                                  &         & \textbf{subword}           & \textbf{word}        & \textbf{span}              & \textbf{document}          & \textbf{order}             & \textbf{$\bm{2\times}$}            & \textbf{$\bm{0.5\times}$}             & \textbf{$\bm{0.25\times}$} \\ \midrule
BoolQ & accuracy & 75.16$^{\pm0.48}$ & 74.87$^{\pm0.26}$ & 75.94$^{\pm0.16}$ & 75.08$^{\pm0.94}$ & 74.75$^{\pm0.71}$ & 74.80$^{\pm1.07}$ & 74.87$^{\pm0.62}$ & 74.84$^{\pm0.71}$ & 74.08$^{\pm0.56}$ \\ [0.5em]
\multirow{1}{*}{CB}            & accuracy                   & 78.93$^{\pm3.43}$      &    76.06$^{\pm2.40}$       & 84.64$^{\pm3.48}$          & 75.71$^{\pm2.71}$          & 82.86$^{\pm1.60}$          & 83.57$^{\pm1.49}$          & 80.00$^{\pm1.96}$          & 74.28$^{\pm3.91}$          & 77.14$^{\pm3.44}$          \\
                               & F\textsubscript{1}                           & 72.11$^{\pm6.73}$    &  72.78$^{\pm5.17}$         & 80.42$^{\pm4.52}$          & 71.91$^{\pm8.36}$          & 77.78$^{\pm2.31}$          & 80.99$^{\pm3.10}$          & 72.56$^{\pm4.04}$          & 66.73$^{\pm4.23}$          & 80.69$^{\pm3.45}$    \\ [0.5em]
CoLA                           & MCC   & 59.36$^{\pm0.96}$                          & 57.17$^{\pm1.92}$          & 58.28$^{\pm0.59}$          & 58.69$^{\pm1.43}$          & 59.73$^{\pm1.34}$          & 57.91$^{\pm1.51}$          & 57.47$^{\pm1.62}$          & 59.98$^{\pm1.40}$          & 58.30$^{\pm1.15}$        \\ [0.5em]
COPA                           & accuracy                      & 60.40$^{\pm5.03}$ &    59.20$^{\pm2.28}$    & 64.00$^{\pm5.43}$          & 59.40$^{\pm5.03}$          & 72.00$^{\pm1.87}$          & 62.80$^{\pm2.77}$          & 54.20$^{\pm1.96}$          & 58.00$^{\pm3.32}$          & 61.60$^{\pm2.19}$   \\ [0.5em]
\multirow{1}{*}{MNLI}          & matched acc.             & 84.22$^{\pm0.12}$   & 85.14$^{\pm0.16}$          & 84.93$^{\pm0.21}$          & 85.05$^{\pm0.19}$          & 85.21$^{\pm0.25}$          & 84.72$^{\pm0.15}$          & 85.17$^{\pm0.16}$          & 84.40$^{\pm0.29}$          & 83.82$^{\pm0.16}$   \\
                               & mismatched acc.   & 84.00$^{\pm0.05}$       & 84.78$^{\pm0.17}$          & 85.05$^{\pm0.13}$          & 85.35$^{\pm0.15}$          & 85.36$^{\pm0.21}$          & 84.73$^{\pm0.19}$          & 85.29$^{\pm0.14}$          & 84.60$^{\pm0.16}$          & 83.71$^{\pm0.16}$    \\
                               & HANS acc.             & 62.47$^{\pm1.68}$    & 64.39$^{\pm1.28}$          & 63.75$^{\pm0.76}$          & 65.60$^{\pm0.53}$          & 60.50$^{\pm1.24}$          & 64.16$^{\pm1.86}$          & 65.32$^{\pm1.14}$          & 62.35$^{\pm0.82}$          & 58.63$^{\pm1.35}$   \\ [0.5em]
\multirow{1}{*}{MRPC}          & accuracy                    & 84.31$^{\pm0.71}$       & 85.00$^{\pm0.94}$          & 85.54$^{\pm0.90}$          & 87.45$^{\pm0.86}$          & 86.52$^{\pm0.81}$          & 85.93$^{\pm0.59}$          & 86.47$^{\pm0.80}$          & 86.27$^{\pm1.12}$          & 85.29$^{\pm0.83}$        \\
                               & F\textsubscript{1} & 89.06$^{\pm0.48}$                           & 89.51$^{\pm0.64}$          & 89.83$^{\pm0.61}$          & 91.20$^{\pm0.62}$          & 90.39$^{\pm0.59}$          & 89.99$^{\pm0.48}$          & 90.54$^{\pm0.61}$          & 90.39$^{\pm0.73}$          & 89.57$^{\pm0.63}$ \\ [0.5em]
MultiRC & F\textsubscript{1} & 67.25$^{\pm0.57}$ & 67.61$^{\pm0.86}$ & 68.10$^{\pm0.85}$ & 71.93$^{\pm0.73}$ & 71.90$^{\pm0.35}$ & 71.91$^{\pm0.35}$ & 66.45$^{\pm2.12}$ & 67.30$^{\pm0.62}$ & 65.02$^{\pm1.00}$ \\
        & exact match & 18.51$^{\pm0.88}$ & 19.58$^{\pm1.51}$ & 18.76$^{\pm1.54}$ & 25.25$^{\pm1.37}$ & 24.91$^{\pm0.40}$ & 27.63$^{\pm0.83}$ & 17.19$^{\pm2.70}$ & 18.65$^{\pm0.44}$ & 16.66$^{\pm0.77}$  \\ [0.5em]

QNLI                           & accuracy            & 90.80$^{\pm0.25}$    & 90.00$^{\pm0.25}$          & 90.57$^{\pm0.29}$          & 91.46$^{\pm0.20}$          & 90.32$^{\pm0.18}$          & 90.36$^{\pm0.25}$          & 90.33$^{\pm0.27}$          & 90.36$^{\pm0.16}$          & 89.08$^{\pm0.24}$  \\ [0.5em]
\multirow{1}{*}{QPP}           & accuracy             & 91.01$^{\pm0.05}$ & 90.94$^{\pm0.06}$          & 90.85$^{\pm0.07}$          & 91.01$^{\pm0.10}$          & 91.00$^{\pm0.14}$          & 90.90$^{\pm0.08}$          & 91.01$^{\pm0.08}$          & 90.77$^{\pm0.04}$          & 90.51$^{\pm0.09}$  \\
                               & F\textsubscript{1} & 87.85$^{\pm0.07}$           & 87.81$^{\pm0.08}$          & 87.73$^{\pm0.07}$          & 87.87$^{\pm0.14}$          & 87.94$^{\pm0.19}$          & 87.76$^{\pm0.10}$          & 87.92$^{\pm0.13}$          & 87.57$^{\pm0.05}$          & 87.24$^{\pm0.13}$    \\ [0.5em]
SST-2                          & accuracy    & 92.06$^{\pm0.48}$         & 92.71$^{\pm0.40}$          & 92.71$^{\pm0.24}$          & 92.80$^{\pm0.50}$          & 92.18$^{\pm0.38}$          & 92.11$^{\pm0.25}$          & 92.34$^{\pm0.59}$          & 92.82$^{\pm0.40}$          & 91.67$^{\pm0.37}$     \\ [0.5em]
\multirow{1}{*}{STS-B}         & Pearson corr.       & 86.34$^{\pm0.29}$  & 87.44$^{\pm0.33}$          & 87.53$^{\pm0.19}$          & 87.99$^{\pm0.11}$          & 89.50$^{\pm0.14}$          & 89.11$^{\pm0.25}$          & 87.83$^{\pm0.19}$          & 86.93$^{\pm0.50}$          & 85.80$^{\pm0.18}$   \\
                               & Spearman corr.    & 86.10$^{\pm0.31}$  & 87.24$^{\pm0.32}$          & 87.45$^{\pm0.20}$          & 87.72$^{\pm0.10}$          & 89.06$^{\pm0.12}$          & 88.82$^{\pm0.22}$          & 87.67$^{\pm0.21}$          & 86.73$^{\pm0.47}$          & 85.54$^{\pm0.20}$        \\ [0.5em]
ReCoRD  & F\textsubscript{1} & 65.48$^{\pm0.64}$ & 63.15$^{\pm3.19}$ & 68.36$^{\pm1.59}$ & 70.71$^{\pm1.81}$ & 66.51$^{\pm0.33}$ & 67.73$^{\pm1.00}$ & 62.93$^{\pm3.12}$ & 64.68$^{\pm1.90}$ & 57.59$^{\pm2.06}$ \\
        & exact match & 64.81$^{\pm0.62}$ & 62.48$^{\pm3.19}$ & 67.61$^{\pm1.58}$ & 70.03$^{\pm1.78}$ & 65.84$^{\pm0.33}$ & 67.04$^{\pm1.02}$ & 62.26$^{\pm3.08}$ & 63.93$^{\pm1.89}$ & 56.88$^{\pm2.07}$ \\ [0.5em]
RTE                            & accuracy               & 62.38$^{\pm3.00}$      & 60.65$^{\pm1.92}$          & 60.51$^{\pm2.07}$          & 60.51$^{\pm2.61}$          & 66.50$^{\pm1.12}$          & 69.68$^{\pm1.28}$          & 58.34$^{\pm2.56}$          & 56.82$^{\pm1.63}$          & 58.19$^{\pm0.59}$         \\ [0.5em]
WiC                            & accuracy               & 66.36$^{\pm1.59}$  &   66.46$^{\pm1.21}$   & 67.40$^{\pm0.43}$          & 69.18$^{\pm1.04}$          & 70.78$^{\pm0.94}$          & 68.90$^{\pm0.60}$          & 67.52$^{\pm1.35}$          & 66.71$^{\pm0.99}$          & 68.46$^{\pm0.71}$           \\ [0.25em] \midrule
\multicolumn{2}{@{}l}{\textbf{Average}}                           & \textbf{74.04$^{\pm2.20}$} & \textbf{73.63$^{\pm1.75}$} & \textbf{75.20$^{\pm1.99}$} & \textbf{75.12$^{\pm2.39}$} & \textbf{76.69$^{\pm0.96}$} & \textbf{76.21$^{\pm1.22}$} & \textbf{73.34$^{\pm1.91}$} & \textbf{73.39$^{\pm1.67}$} & \textbf{73.15$^{\pm1.33}$} \\ \bottomrule
\end{tabular}%
}
\caption{Detailed development GLUE and SuperGLUE results for all tested models. We show the mean and standard deviation statistics over 5 runs with different random seeds (changed only for fine-tuning, the pre-trained models are kept the same).}
\label{tab:finegrained-glue}
\end{table*}

\begin{table*}
\centering
\resizebox{\textwidth}{!}{%
\begin{tabular}{@{}ll@{\hspace{2em}}lllll@{\hspace{2em}}l@{}}
\toprule
\multicolumn{2}{@{}l}{\textbf{Model}}                 & \textbf{POS}               & \textbf{DP}                & \textbf{SRL}               & \textbf{NER}               & \textbf{CR}                & \textbf{Average}           \\ \midrule
\multicolumn{2}{@{}l}{BERT (100M subset)}   & 97.94$^{\pm0.01}$ & 95.03$^{\pm0.04}$ & 92.34$^{\pm0.06}$ & 95.91$^{\pm0.12}$ & 95.27$^{\pm0.10}$ & 95.30$^{\pm0.08}$ \\ \midrule
\multirow{3}{*}{MLM}      & subword       & 97.91$^{\pm0.01}$ & 94.99$^{\pm0.02}$ & 92.44$^{\pm0.03}$ & 95.77$^{\pm0.06}$ & 95.30$^{\pm0.07}$ & 95.28$^{\pm0.05}$ \\
                          & whole-word    & 97.90$^{\pm0.01}$ & 94.99$^{\pm0.05}$ & 92.42$^{\pm0.08}$ & 95.71$^{\pm0.07}$ & 95.64$^{\pm0.07}$ & 95.33$^{\pm0.06}$ \\
                          & span          & 97.91$^{\pm0.01}$ & 94.80$^{\pm0.03}$ & 92.32$^{\pm0.02}$ & 95.56$^{\pm0.07}$ & 95.46$^{\pm0.14}$ & 95.21$^{\pm0.07}$ \\ \midrule
\multirow{2}{*}{NSP}      & subword + document      & 97.92$^{\pm0.01}$ & 95.01$^{\pm0.03}$ & 92.42$^{\pm0.06}$ & 95.76$^{\pm0.07}$ & 95.25$^{\pm0.11}$ & 95.28$^{\pm0.07}$ \\
                          & subword + order         & 97.85$^{\pm0.01}$ & 94.92$^{\pm0.06}$ & 92.25$^{\pm0.07}$ & 95.22$^{\pm0.05}$ & 95.25$^{\pm0.11}$ & 95.10$^{\pm0.07}$ \\ \midrule
\multirow{3}{*}{Steps}     & subword + $2\times$        & 97.93$^{\pm0.01}$ & 94.95$^{\pm0.10}$ & 92.47$^{\pm0.03}$ & 95.63$^{\pm0.11}$ & 95.58$^{\pm0.04}$ & 95.31$^{\pm0.07}$ \\
                          & subword + $\nicefrac{1}{2}\times$         & 97.90$^{\pm0.02}$ & 95.02$^{\pm0.04}$ & 92.38$^{\pm0.05}$ & 95.46$^{\pm0.03}$ & 95.43$^{\pm0.05}$ & 95.24$^{\pm0.04}$ \\
                          & subword + $\nicefrac{1}{4}\times$          & 97.88$^{\pm0.01}$ & 94.81$^{\pm0.07}$ & 92.21$^{\pm0.03}$ & 95.32$^{\pm0.08}$ & 95.00$^{\pm0.18}$ & 95.04$^{\pm0.10}$ \\ \midrule
\multicolumn{2}{@{}l}{Random initialization} & 69.85$^{\pm0.42}$ & 66.25$^{\pm0.20}$ & 70.87$^{\pm0.21}$ & 73.16$^{\pm0.60}$ & 85.56$^{\pm0.46}$ & 73.14$^{\pm0.41}$ \\ \bottomrule
\end{tabular}%
}
\caption{Detailed edge probing results for all tested models. ${}^\dagger$ The $\textrm{BERT}_\textit{base}$ scores in the first row are taken from \newcite{tenney2018what}. The last row shows the edge probing results with a randomly initialized language model -- its performance hints at how much information is included in the probes themselves.}
\label{tab:edge-probing-finegrained}
\end{table*}

\begin{table*}
\centering
\resizebox{\textwidth}{!}{%
\begin{tabular}{@{}lr@{\hspace{2em}}ccc@{\hspace{2em}}cc@{\hspace{2em}}ccc@{}}
\toprule
\multirow{2}{*}{\textbf{BLiMP subgroups}} & \multirow{2}{*}{\hspace{-2.5em}\textbf{BERT (100M subset)}} & \multicolumn{3}{c@{\hspace{2em}}}{\crulefill~~\textbf{MLM}~~\crulefill}                       & \multicolumn{2}{c@{\hspace{2em}}}{\crulefill~~\textbf{NSP}~~\crulefill}   & \multicolumn{3}{c}{\crulefill~~\textbf{Size}~~\crulefill}                \\
                                          &                                    & \textbf{subword} & \textbf{word} & \textbf{span} & \textbf{document} & \textbf{order} & \textbf{medium} & \textbf{small} & \textbf{tiny} \\ \midrule
Anaphor agreement                        & 93.20                              & 93.95            & 92.65 & 94.50               & 93.00         & 94.00             & 94.65          & 93.30           & 94.60          \\
Argument structure                       & 78.95                              & 80.73            & 67.93 & 80.98               & 81.58         & 80.61             & 81.54          & 80.98           & 81.99    \\
Binding                                   & 77.04                              & 78.34            & 74.60 & 77.26               & 77.74         & 76.60             & 77.33          & 76.43           & 77.03       \\
Control/raising                          & 73.76                              & 79.68            & 79.90 & 81.02               & 78.18         & 78.32             & 78.84          & 79.80           & 78.82        \\
Determiner-noun agreement               & 95.91                              & 96.74            & 93.48 & 97.45               & 97.09         & 96.26             & 97.09          & 96.73           & 96.96          \\
Ellipsis                                  & 88.25                               & 88.10            & 85.55 & 90.95               & 88.65         & 86.70             & 90.25          & 87.70           & 88.70          \\
Filler-gap                 & 85.44                              & 83.87        & 83.30    & 85.86               & 87.23         & 83.46             & 85.20          & 84.73           & 84.20       \\
Irregular forms                          & 88.45                               & 91.45            & 86.75 & 94.40               & 88.30         & 86.70             & 92.35          & 92.65           & 93.10       \\
Island effects                           & 70.91                              & 74.99            & 76.71 & 74.34               & 73.98         & 74.62            & 72.14          & 74.86           & 72.34       \\
NPI licensing                            & 81.07                              & 82.40            & 81.73 & 82.36               & 83.24         & 78.79             & 82.43          & 84.96           & 82.86      \\
Quantifiers                               & 69.98                              & 68.88            & 70.00 & 74.13              & 64.50         & 68.77             & 72.58          & 67.10           & 67.53    \\
Subject-verb agreement                  & 91.78                               & 92.64            & 83.97 & 92.92               & 92.13         & 90.00             & 91.72          & 92.25           & 92.22     \\ \midrule
\textbf{Accuracy} & \textbf{81.95} & \textbf{83.42} & \textbf{80.05} & \textbf{84.18} & \textbf{83.31} & \textbf{82.17} & \textbf{83.45} & \textbf{83.47} & \textbf{83.15} \\\bottomrule
\end{tabular}%
}
\caption{Detailed BLiMP results for all tested models.}
\label{tab:blimp-detailed}
\end{table*}

\begin{table*}
\centering
\begin{tabular}{@{}lcccc@{}}
\toprule
\textbf{Hyperparameter} & \textbf{Base} \\ \midrule
Number of layers        & 12            \\
Hidden size             & 768           \\
FF intermediate size    & 2\,048        \\
Vocabulary size         & 16\,384          \\
FF activation function  & GEGLU         \\
Attention heads         & 12            \\
Attention head size     & 64            \\
Dropout                 & 0.1           \\
Attention dropout       & 0.1           \\
Training steps          & 31\,250       \\
Batch size              & 32\,768 (90\% steps) / 8\,192 (10\% steps)        \\
Sequence length         & 128 (90\% steps) / 512 (10\% steps)         \\
Tokens per step         & 4\,194\,304   \\
Warmup steps            & 500 (1.6\% steps)         \\
Initial learning rate   & 0.01          \\
Final learning rate     & 0.001          \\
Learning rate decay     & cosine        \\
Weight decay            & 0.1          \\
Layer norm $\epsilon$   & 1e-5          \\
Optimizer               & LAMB         \\
LAMB $\epsilon$         & 1e-6          \\
LAMB $\beta_1$          & 0.9           \\
LAMB $\beta_2$          & 0.98          \\
Gradient clipping       & 2.0           \\ \bottomrule
\end{tabular}
\caption{Pre-training hyperparameters. The models differ only in their hidden size and number of layers, the learning rate schedule and other training settings are kept identical.}
\label{tab:hyperparams}
\end{table*}

\begin{table*}
\centering
\resizebox{\textwidth}{!}{%
\begin{tabular}{@{}lccccc@{}}
\toprule
\multirow{3}{*}{\textbf{Hyperparameter}} & \multirow{3}{*}{\textbf{ReCoRD}} & \multirow{3}{*}{\textbf{MNLI, QQP, QNLI}} & \textbf{BoolQ, CoLA, COPA,} & \multirow{3}{*}{\textbf{RTE, WiC}} & \multirow{3}{*}{\textbf{CB}} \\
& & & \textbf{SST-2, MultiRC,} & & \\
& & & \textbf{MRPC, STSB} & & \\ \midrule

Batch size             & 32 & 32                       & 32                                 & 16                & 8           \\
Number of epochs       & 1 & 4                        & 8                                                           & 8                 & 16          \\
Dropout              & 0.1      & 0.1                      & 0.1                                                         & 0.1               & 0.1         \\
Warmup steps          & 10\%     & 10\%                     & 10\%                                                        & 10\%              & 10\%        \\
Peak learning rate    & 3e-5     & 3e-5                     & 3e-5                                                        & 3e-5              & 3e-5        \\
Learning rate decay  & linear    & linear                   & linear                                                      & linear            & linear      \\
Weight decay           & 0.01   & 0.01                     & 0.01                                                        & 0.01              & 0.01        \\
Optimizer             & AdamW        & AdamW                    & AdamW                                                       & AdamW             & AdamW       \\
Adam $\epsilon$     & 1e-6    & 1e-6                     & 1e-6                                                        & 1e-6              & 1e-6        \\
Adam $\beta_1$       & 0.9        & 0.9                      & 0.9                                                         & 0.9               & 0.9         \\
Adam $\beta_2$     & 0.999        & 0.999                    & 0.999                                                       & 0.999             & 0.999       \\ \bottomrule
\end{tabular}%
}
\caption{Hyperparameters for fine-tuning the GLUE and SuperGLUE tasks. We use the same hyperparamaters for all models, not performing any per-model hyperparameter search.}
\label{tab:glue-hyperparams}
\end{table*}

\begin{table*}
\centering
\begin{tabular}{@{}lcc@{}}
\toprule
\textbf{Hyperparameter} & \textbf{POS, SRL, NER, CR} & \textbf{DP} \\ \midrule
Batch size              & 128                           & 128         \\
Number of epochs        & 5                             & 10                               \\
Dropout                 & 0.25                          & 0.25                             \\
Downsampled hidden size & 256 & 256 \\
Attention pooling heads & 4 & 4 \\
MLP hidden layers       & 1 & 1 \\
Starting learning rate  & 6e-3                          & 6e-3                             \\
Learning rate decay     & cosine                        & cosine                           \\
Weight decay            & 0.01                          & 0.01                             \\
Optimizer               & AdamW                         & AdamW                            \\
Adam $\epsilon$         & 1e-6                          & 1e-6                             \\
Adam $\beta_1$          & 0.9                           & 0.9                              \\
Adam $\beta_2$          & 0.999                         & 0.999                            \\
Gradient clipping       & 2.0                           & 2.0         \\ \bottomrule
\end{tabular}
\caption{Edge probing hyperparameters.}
\label{tab:ep-hyperparams}
\end{table*}




\begin{listing*}[ht]
\begin{minted}[xleftmargin=1.7em, linenos=true, breaklines, breakafter=d, fontsize=\scriptsize]{markdown}
# Oral history project: interview

Britta: 'Can you tell us what er what section you work in?'

Eliazar: 'I work at the weaving'

Britta: 'In the weaving?'

Eliazar: 'section, aha.
And'

Britta: 'And what do you do?'

Eliazar: 'I'm what you call a Axminster handler'

Britta: 'Aha.'

Eliazar: 'which involves like when the frames comes off the weaving and they're yarn left, I strip the yarn off.'

Britta: 'Mhm.'

Eliazar: 'Off the, the, the weaving frames.'

Britta: 'Mhm.'

Eliazar: 'That's basically my, aha.'

Britta: 'It's quite spec specialized so'

Eliazar: 'No no, no, no.
It's not specialized, no.'

Britta: 'Mhm, have you ever worked in any other factory?'

Eliazar: 'Aha, I worked in spooling, I've been left now two year.'

Britta: 'And how did you find that?'

Eliazar: 'Er, I liked the spooling but some I just don't know, some of the girls get kind of one [UNK] one thing by the other I can object to, I think it was actually the atmosphere of the, the girls that worked in the department that I'
\end{minted}
\caption{A random example of the first few lines from a preprocessed spoken document from BNC. Notice that the test is divided into speech turns (paragraphs), each starting with the name of a speaker. Line 39 contains a special \texttt{[UNK]} token in place of an incomprehensible word or phrase.}
\label{list:spoken}
\end{listing*}

\begin{listing*}[ht]
\begin{minted}[xleftmargin=1.7em, linenos=true, breaklines, breakafter=|, fontsize=\scriptsize]{markdown}
# Organizing knowledge: an introduction to information retrieval

## SUBJECTS

### The subject approach: introduction, processes, tools and simple evaluation

#### 1.2.1 Subjects

Users often approach information sources not with names (as have been considered in Part II), but with a question that requires an answer or a topic for study.
Users seek documents or information concerned with a particular subject.
In order to make some provision for this common approach to information sources, it is necessary to arrange documents- and document surrogates in catalogues, indexes bibliographies, computer databases and so on - in such a way that items on specific subjects can be retrieved.
Thus, the subject approach is extremely important in the access to and the exploitation of information, documents and data.

Before we discuss the provision that libraries and information workers make for the subject approach, it may be useful to consider the preliminary question: What is a subject?
In talking about a subject we generally refer to a given area of knowledge or to the contents of an information source of a given scope.
A subject might be considered to be defined by:

- an area of interest,

- an area in which an individual researcher or professional works,

- an area in which an individual writes or an area of knowledge which is studied.
\end{minted}
\caption{A sample of the first few lines from a written BNC article. Note the H1-level header with the title of the whole document and then the title of a chapter, section and subsection in the lines below. This sample also contains a special text block with a list in the last lines.}
\label{list:written}
\end{listing*}

\end{document}